# Self-Supervised Implicit CEST Reconstruction via Physics-Informed Lorentz Encoding

Dexuan Li[#], Yupeng Wu[#], Chenglong Wang, Hanlin Liu, Hui Zheng, Jianqi Li, and Guang Yang[*]

Shanghai Key Laboratory of Magnetic Resonance, Institute of Magnetic Resonance and Molecular Imaging in Medicine, East China Normal University
gyang@phy.ecnu.edu.cn

**Abstract.** Multi-Pool Chemical Exchange Saturation Transfer (CEST) MRI provides valuable metabolic information but is clinically limited by long acquisition times. Although sparse sampling reduces scanning time, reconstructing high-resolution Z-spectra from limited data remains an ill-posed inverse problem. Conventional interpolation and generic Implicit Neural Representations (INRs) often **lack physical constraints**, leading to spectral artifacts and physically invalid signals. To address this, we propose **Lorentz Encoding (LE)**, a physics-informed framework that formulates CEST reconstruction as a self-supervised reconstruction task via implicit continuous coordinate learning. Unlike generic positional encodings, LE regularizes the continuous spectral mapping by projecting sparse coordinates into a physically constrained space governed by a combination of parametric Lorentzian profiles with learnable basis functions. This mechanism effectively reduces noise and enforces consistency with physical models. Experiments on in vivo human brain data demonstrate that LE significantly outperforms state-of-the-art methods. Specifically, under a 39-point sampling strategy, LE achieves a PSNR of **57.58 dB** and an SSIM of **0.9994**. Furthermore, the learned physics-informed encodings form a continuous, geometrically ordered trajectory in the latent space, ensuring accurate quantitative metabolite mapping (APT, NOE, MT).

## 1 Introduction

Multi-Pool Chemical Exchange Saturation Transfer (CEST) MRI has emerged as a vital tool for molecular imaging, enabling the non-invasive detection of low-concentration endogenous metabolites[1]. Its diagnostic capability relies on the Z-spectra to quantify molecular effects such as Amide Proton Transfer (APT), Nuclear Overhauser enhancement (NOE) effect, and Magnetization Transfer (MT). Fundamentally, CEST exploits frequency-selective radiofrequency (RF) irradiation to saturate dilute exchangeable solute protons; through continuous chemical exchange, this saturation state is transferred to the abundant bulk water pool, resulting in an amplified attenuation of the water signal that is recorded as a function of the frequency offset to form the Z-spectrum[1,2]. However, capturing these subtle spectral features typically requires dense sampling (e.g., >50 frequency offsets), leading to long acquisition times that severely limit clinical translation[3].

While sparse sampling offers a method to accelerate scanning, it renders the Z-spectra reconstruction an ill-posed inverse problem. Traditional model-based approaches such as Multi-Pool Lorentzian Fitting (MPLF) are theoretically sound but numerically unstable when data are sparse. Recently, deep learning approaches such as variants of U-Net[4, 5] and generic Implicit Neural Representation (INR)[6–8] have demonstrated the capability to model continuous spectral signals from sparse data. In the INR paradigm, the reconstruction of 2D CEST is typically formulated as learning a continuous function mapping spatial-spectral coordinates $(x, y, \Delta\omega)$ to the normalized Z-spectra intensity. However, standard INRs utilizing generic encodings (e.g., Fourier[9] or Hash Encoding[10]) treat all dimensions equally and lack physical constraints. By treating the Z-spectra as an arbitrary continuous function, they force the network to search for solutions in an unconstrained high-dimensional space. Consequently, they often overfit high-frequency noise or produce spectral oscillations that are physically invalid.

To address this, we propose Lorentz Encoding (LE), a physics-informed framework that fundamentally reformulates CEST reconstruction from unconstrained regression to physical-constrained learning. Unlike generic methods, our approach explicitly decouples the spatial and spectral dimensions to accommodate their distinct natures: spatial features are textural and high frequency, while spectral features are smooth, governed by physical laws. For the spectral dimension of CEST images, LE embeds Lorentzian priors directly into the encoding layer. By forcing the continuous spatial-spectral mapping to be constructed from a combination of these physically valid bases, our method restricts the solution space to biological signals. This physical-informed constraint effectively filters out non-physical high-frequency fluctuations and speckle artifacts, enabling robust reconstruction and accurate quantitative parameter mapping (APT, NOE, MT) from as few as 21 sampling points.

Our main contributions are summarized as follows: 1) We propose **Lorentz Encoding (LE)** within an INR framework, transforming sparse CEST reconstruction from unconstrained regression into a strictly self-supervised, physics-informed optimization paradigm. 2) We introduce a **decoupled hybrid architecture** that utilizes Hash Encoding for high-frequency spatial textures and LE for physical spectral prior injection, effectively eliminating artifacts. 3) Extensive in-vivo experiments demonstrate that LE significantly outperforms SOTA baselines under extreme acceleration (e.g., 21 points), ensuring accurate downstream metabolite mapping (APT, NOE, MT).

# 2 Method

## 2.1 Overall Architecture

The goal of our framework is to reconstruct dense, high-fidelity CEST Z-spectra from highly under-sampled acquisitions. Unlike traditional discrete grid representations, INR models a continuous signal by mapping spatial or spectral coordinates directly to signal intensities. Thus, we formulate the reconstruction of 2D CEST as learning a con-tinuous mapping function $F_\Theta: \mathbb{R}^3 \to \mathbb{R}$, which translates a given spatial-spectral coordinate $(x, y, \Delta\omega)$ into a normalized signal intensity $\hat{S}$. Physically, a Z-spectrum is essentially a superposition of multiple Lorentzian line shapes. To exploit this prior and prevent the network from blindly searching in an unconstrained space, we propose a decoupled hybrid encoding architecture. As illustrated in **Fig. 1**, to accommodate the distinct inductive biases required for different dimensions, we propose a decoupled hybrid encoding architecture before the Multi-Layer Perceptron (MLP) network.

The forward process is mathematically defined through the two parallel encoding modules. First, the spatial coordinates $(x, y)$ are encoded using Multi-resolution Hash Encoding (MHE)[10] to capture high-frequency anatomical details, denoted as $\Phi_{spa}(x, y)$. Concurrently, the spectral offset $\Delta\omega$ is projected onto a continuous, physics-informed spectral encoding space via our proposed Lorentz Encoding (LE). This yields $\Phi_{spec}(\Delta\omega) = [L_1(\Delta\omega), \ldots, L_N(\Delta\omega)]^T$, where $\{L_i\}_{i=1}^N$ represents a set of learnable Lorentzian basis functions that jointly define a valid physical subspace in $\mathbb{R}^N$.

The decoupled features are then concatenated and decoded by an MLP network $\Theta$:

$$\hat{S} = F_\Theta(x, y, \Delta\omega) = MLP_\Theta\left(\Phi_{spa}(x, y) \oplus \Phi_{spec}(\Delta\omega)\right) \tag{1}$$

where $\oplus$ denotes the channel-wise concatenation. The entire network is optimized end-to-end by minimizing the $L_2$ error $\mathcal{L} = \sum_\Omega |\hat{S} - \hat{S}_{gt}|_2^2$ with ground-truth signals $\hat{S}_{gt}$ over the sparse acquisition set $\Omega$ subject to the physical bounds defined in **Table 1**. This hybrid design keeps spatial details while strictly confining the spectral interpolation to a physically plausible, noise-resilient representation space.

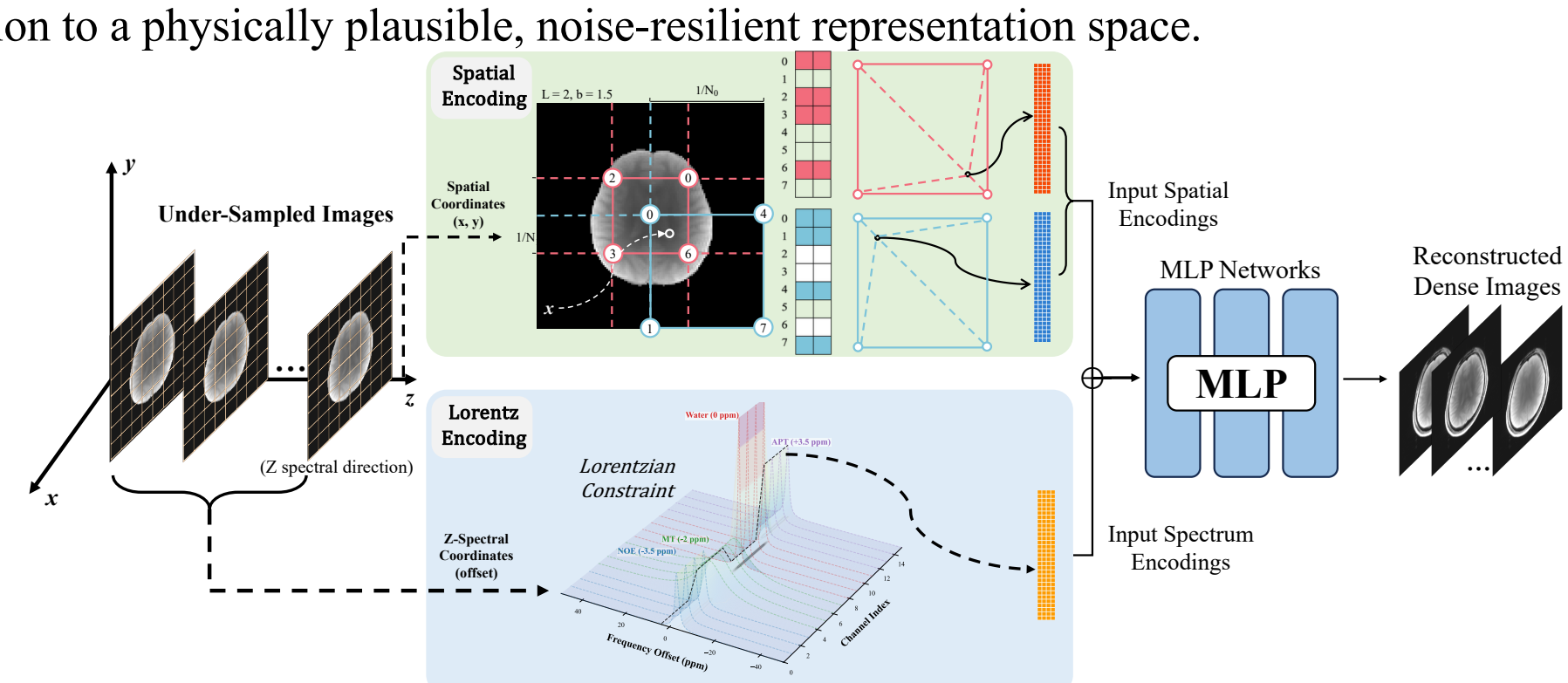

**Fig. 1.** Overview of LE-INR workflow.

## 2.2 Lorentz Encoding: Physical Prior Injection

**Core Formulation.** Standard positional encoding is used to map spatial coordinates to feature spaces, while LE projects spectral coordinates onto a parameterized physical subspace spanned by a set of $N$ learnable basis functions.

Formally, for a given frequency offset $\Delta\omega$, the encoding vector $L_i(\Delta\omega)$ is computed as:

$$L_i(\Delta\omega) = A_i \frac{\gamma_i^2/4}{\gamma_i^2/4 + (\Delta\omega - \mu_i)^2} \tag{2}$$

where each basis function $L_i$ represents a distinct Lorentzian pool parameterized by learnable components: amplitude $A_i$,center frequency $\mu_i$, and peak width $\gamma_i$. Restricting the MLP input to this basis combination transforms the task into physical-constrained optimization, naturally filtering out noise and artifacts that deviate from the Lorentzian line shape.

**Physics-Informed Structured Initialization.** To enforce physical plausibility, we allocate the $N$-dimensional encoding subspace equally among four distinct solute pools: Water, APT, NOE, and MT. The learnable parameters for each distinct Lorentzian pool—amplitude ($A_i$), center frequency ($\mu_i$), and peak width ($\gamma_i$)—are uniformly initialized and strictly bounded during optimization within physically determined bounds detailed in **Table 1**.

**Table 1.** Physically determined parameter ranges for structured initialization.

| Pool | Center frequency $\mu_i$ $(ppm)$ | Peak Width $\gamma_i$ $(ppm)$ | Amplitude $A_i$ |
|---|---|---|---|
| **Water** | [-0.5, 0.5] | [1, 2] | [0.5, 1.0] |
| **APT** | [3.0, 4.0] | [3, 4] | [0, 0.5] |
| **NOE** | [-4.0, -3.0] | [4, 5] | [0, 0.5] |
| **MT** | [-1.5, -0.5] | [25, 30] | [0, 0.5] |

## 2.3 Network Configuration

All models were implemented using the PyTorch 2.8 framework and trained on a single NVIDIA RTX 5080 GPU (16GB VRAM). The implicit neural representation consists of a Multi-Layer Perceptron (MLP) with 6 fully connected layers, each containing 128 hidden units followed by ReLU activation. The output layer also uses a ReLU activation to regress the normalized signal intensity.

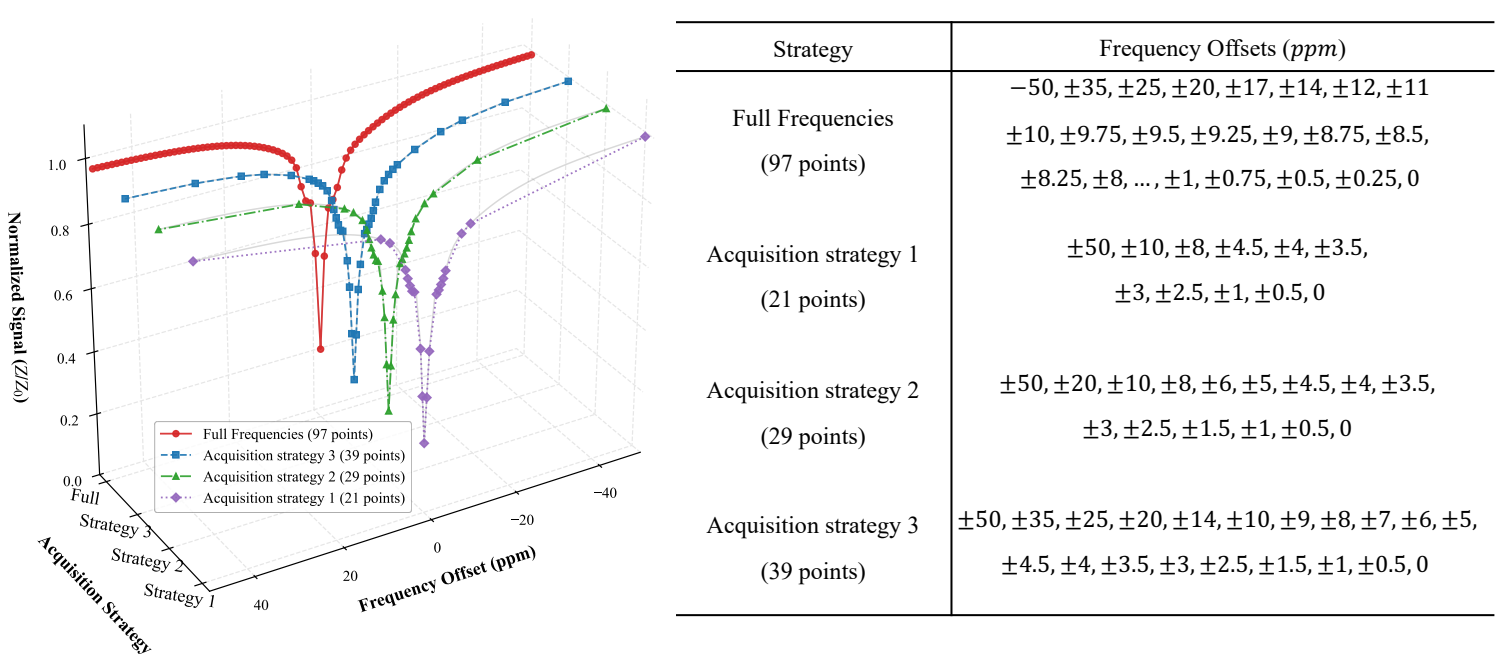

| Strategy | Frequency Offsets $(ppm)$ |
|---|---|
| Full Frequencies (97 points) | $-50, \pm35, \pm25, \pm20, \pm17, \pm14, \pm12, \pm11$ $\pm10, \pm9.75, \pm9.5, \pm9.25, \pm9, \pm8.75, \pm8.5,$ $\pm8.25, \pm8, \dots, \pm1, \pm0.75, \pm0.5, \pm0.25, 0$ |
| Acquisition strategy 1 (21 points) | $\pm50, \pm10, \pm8, \pm4.5, \pm4, \pm3.5,$ $\pm3, \pm2.5, \pm1, \pm0.5, 0$ |
| Acquisition strategy 2 (29 points) | $\pm50, \pm20, \pm10, \pm8, \pm6, \pm5, \pm4.5, \pm4, \pm3.5,$ $\pm3, \pm2.5, \pm1.5, \pm1, \pm0.5, 0$ |
| Acquisition strategy 3 (39 points) | $\pm50, \pm35, \pm25, \pm20, \pm14, \pm10, \pm9, \pm8, \pm7, \pm6, \pm5,$ $\pm4.5, \pm4, \pm3.5, \pm3, \pm2.5, \pm1.5, \pm1, \pm0.5, 0$ |

(a) 3D visualization of sampled signals. (b) Exact frequency offsets for each strategy.

**Fig. 2.** Acquisition Strategy of Full-Sampled and Under-sampled Z-spectra

# 3 Experiments and Results

## 3.1 Materials

Eighteen healthy subjects were enrolled in this study (IRB approved, written consent obtained). We acquired a reference 97-offset Z-spectrum (~9 min) on a 3T Siemens Prisma using a 2D single-shot FLASH sequence ( $TR = 4\,ms$, $\text{FOV} = 256 \times 256\,\text{mm}^2$, $spatial\ resolution = 2 \times 2 \times 10\,mm^3$). Saturation was achieved via 28 Gaussian RF pulses ($0.6\,\mu T, 3\,s$). Following standard B0 inhomogeneity correction (shifting the signal minimum to 0 ppm), the dense data was retrospectively under-sampled using the protocols defined in **Fig. 2**. Our source code is publicly available at **https://github.com/ldxFAIRYTAIL/Lorentz-Encoding.git**

## 3.2 Training and Optimization

Our framework follows the subject-specific optimization paradigm inherent to Implicit Neural Representations (INR). For a given 2D slice ($128 \times 128$), the training set consists exclusively of the coordinate-intensity pairs from the $M$ sparse acquired frequency offsets $\Omega$ (e.g., $128 \times 128 \times 21 = 344{,}064$ pairs under Strategy 1). During each epoch, 10,000 points are randomly sampled from $\Omega$. The Adam optimizer jointly updates the network parameters $\Theta$ and the learnable physical constants $(A_i, \mu_i, \gamma_i)$ of the Lorentzian pools by minimizing the point-wise $L_2$ loss over 10,000 epochs. The remaining unseen frequency offsets $(97 - M)$ serve strictly as the unseen test set to evaluate interpolation fidelity and downstream metabolite mapping.

## 3.3 Results

We benchmark LE against Direct MPLF (Direct MPLF does not perform any interpolation. It directly fits parameters using only the acquired sparse data points), as well as a CNN baseline and three state-of-the-art INR strategies: **1) U-Net:** A standard 4-stage encoder-decoder (channels: 32, 64, 128, 256) mapping sparse inputs to dense spectra. **2) WIRE[11]:** An implicit representational strategy utilizing complex Gabor activations. **3) Fourier PE[9]:** Explicit positional encoding with $N_{freq} = 64$. **4) MHE[10]:** Multi-resolution Hash Encoding with $n_{levels} = 16$. For our proposed LE, we utilize $N = 16$ learnable Lorentzian bases. Serving as a direct ablation study, MHE replaces our 16-dimensional LE with a 16-dimensional 1D-MHE for the spectral axis. Both methods concatenate 32-dimensional spatial and 16-dimensional spectral features, maintaining identical network capacity for a fair comparison.

**Z-spectra Reconstruction Quality. Table 2** summarizes the quantitative evaluation of Z-spectra reconstruction across different sampling strategies. As shown, our proposed LE consistently outperformed all baselines across all sampling strategies. **Specifically**, under Strategy 3 (39 points), LE achieved the best performance with a PSNR of **57.58 dB** and SSIM of **0.9994**. Even with the most aggressive under-sampling (Strategy 1, 21 points), LE maintained a high PSNR of 54.19 dB, significantly higher than U-Net (42.98 dB) and MHE (33.56 dB).

**Quantitative Parameter Mapping.** To assess clinical utility, we derived quantitative APT, NOE, and MT maps. Visualizations for a representative case are shown in

**Fig. 3**. For all deep learning-based methods, these maps were generated by applying standard MPLF to their respectively reconstructed dense Z-spectra. As shown in **Table 2**, parameter maps derived from LE reconstruction consistently yielded the highest PSNR and SSIM relative to the reference. For instance, for MT maps under Strategy 3, LE achieved a PSNR of 32.69 dB, ensuring highly reliable quantitative mapping.

**Table 2.** Metrics Results of Image Reconstruction and Parameter Maps

| Strategy | Method | Metrics of Reconstructed Z-spectra (PSNR/SSIM) | Metrics of Parameter APT Map (PSNR/SSIM) | Metrics of Parameter MT Map (PSNR/SSIM) | Metrics of Parameter NOE Map (PSNR/SSIM) |
|---|---|---|---|---|---|
| Acquisition strategy 1 (21 $points$) | Direct MPLF | - | 23.93 ± 1.67<br>0.8176 ± 0.0215 | 21.38 ± 1.51/<br>0.8431 ± 0.0135 | 22.42 ± 2.43/<br>0.7872 ± 0.0237 |
| | U-Net | 42.98 ± 2.71/<br>0.9926 ± 0.0034 | 22.82 ± 1.83/<br>0.7309 ± 0.0235 | 22.18 ± 2.69/<br>0.7747 ± 0.0336 | 23.18 ± 2.34/<br>0.7206 ± 0.0269 |
| | WIRE | 30.28 ± 0.58/<br>0.9588 ± 0.0103 | 18.47 ± 1.51/<br>0.6615 ± 0.0302 | 14.22 ± 1.14/<br>0.6461 ± 0.0217 | 19.72 ± 2.58/<br>0.6905 ± 0.0342 |
| | Fourier PE | 27.95 ± 0.87/<br>0.9667 ± 0.0135 | 16.80 ± 1.41/<br>0.6353 ± 0.0226 | 12.06 ± 1.09/<br>0.6296 ± 0.0196 | 17.96 ± 2.23/<br>0.6537 ± 0.0281 |
| | MHE | 33.56 ± 0.46/<br>0.9874 ± 0.0018 | 19.95 ± 1.55/<br>0.7056 ± 0.0255 | 14.84 ± 0.90/<br>0.6902 ± 0.0174 | 21.79 ± 2.18/<br>0.7326 ± 0.0329 |
| | **LE (ours)** | **54.19 ± 2.06/<br>0.9991 ± 0.0002** | **29.38 ± 1.79/<br>0.8605 ± 0.0259** | **28.02 ± 1.74/<br>0.8922 ± 0.0258** | **28.97 ± 2.64/<br>0.8379 ± 0.0298** |
| Acquisition strategy 2 (29 $points$) | Direct MPLF | - | 28.97 ± 1.67/<br>0.8881 ± 0.0167 | 26.79 ± 1.76/<br>0.9038 ± 0.0129 | 27.33 ± 2.53/<br>0.8649 ± 0.0227 |
| | U-Net | 42.17± 3.38/<br>0.9886 ± 0.0086 | 21.60 ± 2.17/<br>0.7263 ± 0.0249 | 20.07 ± 3.31/<br>0.7680 ± 0.0463 | 21.62 ± 3.22/<br>0.7237 ± 0.0272 |
| | WIRE | 33.49 ± 1.11/<br>0.9756 ± 0.0056 | 21.05 ± 1.38/<br>0.7111 ± 0.0235 | 15.70 ± 0.84/<br>0.6766 ± 0.0187 | 21.35 ± 2.39/<br>0.7363 ± 0.0222 |
| | Fourier PE | 31.29 ± 0.72/<br>0.9713 ± 0.0102 | 20.75 ± 1.26/<br>0.6829 ± 0.0245 | 12.49 ± 1.04/<br>0.6383 ± 0.0182 | 21.46 ± 2.38/<br>0.7222 ± 0.0236 |
| | MHE | 37.72 ± 0.34/<br>0.9948 ± 0.0013 | 23.93 ± 1.48/<br>0.7983 ± 0.0182 | 16.87 ± 0.89/<br>0.7434 ± 0.0184 | 23.96 ± 2.26/<br>0.7973 ± 0.0276 |
| | **LE (ours)** | **56.53 ± 1.35/<br>0.9993 ± 0.0001** | **31.60 ± 1.64/<br>0.9003 ± 0.0149** | **31.01 ± 1.73/<br>0.9338 ± 0.0179** | **31.43 ± 2.60/<br>0.8853 ± 0.0223** |
| Acquisition strategy 3 (39 $points$) | Direct MPLF | - | 32.79 ± 1.65/<br>0.9312 ± 0.0114 | 30.84 ± 1.76/<br>0.9396 ± 0.0108 | 31.71 ± 2.55/<br>0.9110 ± 0.0219 |
| | U-Net | 44.42 ± 2.49/<br>0.9926 ± 0.0040 | 23.25 ± 1.71/<br>0.7686 ± 0.0174 | 21.98 ± 2.17/<br>0.8128 ± 0.0314 | 23.69 ± 2.33/<br>0.7481 ± 0.0314 |
| | WIRE | 41.87 ± 1.83/<br>0.9865 ± 0.0049 | 23.62 ± 1.21/<br>0.7763 ± 0.0268 | 22.61 ± 1.34/<br>0.8055 ± 0.0415 | 23.59 ± 2.75/<br>0.7645 ± 0.0324 |
| | Fourier PE | 44.36 ± 1.08/<br>0.9909 ± 0.0035 | 25.43 ± 1.50/<br>0.8022 ± 0.0172 | 24.23 ± 1.11/<br>0.8392 ± 0.0174 | 25.24 ± 2.31/<br>0.7937 ± 0.0235 |
| | MHE | 48.77 ± 0.57/<br>0.9978 ± 0.0003 | 27.62 ± 1.48/<br>0.8786 ± 0.0145 | 27.30 ± 1.40/<br>0.9099 ± 0.0139 | 27.40 ± 2.10/<br>0.8539 ± 0.0230 |
| | **LE (ours)** | **57.58 ± 1.26/<br>0.9994 ± 0.0001** | **33.04± 1.65/<br>0.9241 ± 0.0140** | **32.69 ± 1.67/<br>0.9503 ± 0.0121** | **32.66 ± 2.37/<br>0.9055 ± 0.0203** |

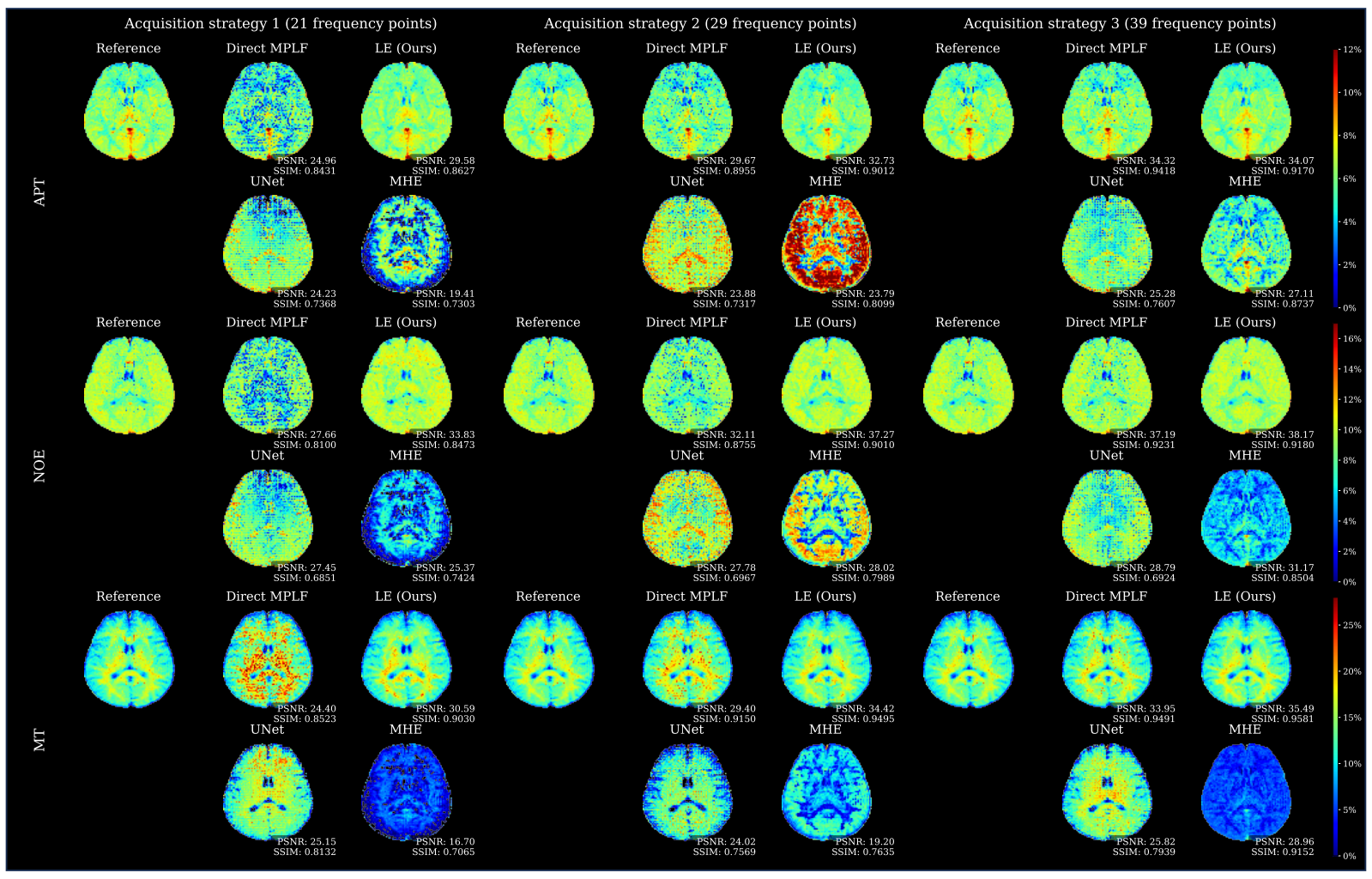

**Fig. 3.** Quantitative parameter mapping for a representative case

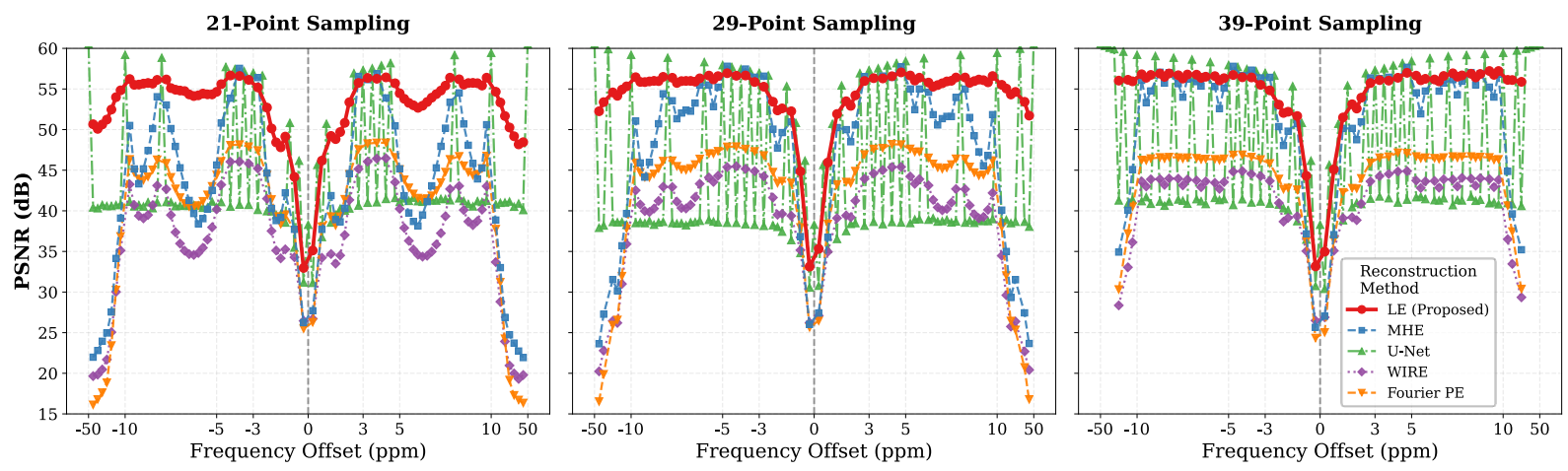

**Fig. 4.** PSNR at different frequency points for three sampling strategies.

**Spectral Fidelity.** We further analyzed the reconstruction quality across different frequency offsets. As shown in **Fig. 4**, which plots the PSNR evaluated at all interpolated sampling points, baseline methods (e.g., WIRE, MHE) exhibit significant performance drops in spectral regions distant from the water peak (0 ppm). In contrast, LE maintains consistent high PSNR across the entire Z-spectra. Furthermore, the learned physical encodings is visualized using Principal Component Analysis (PCA) in **Fig. 5** and discussed in detail in Section 4.

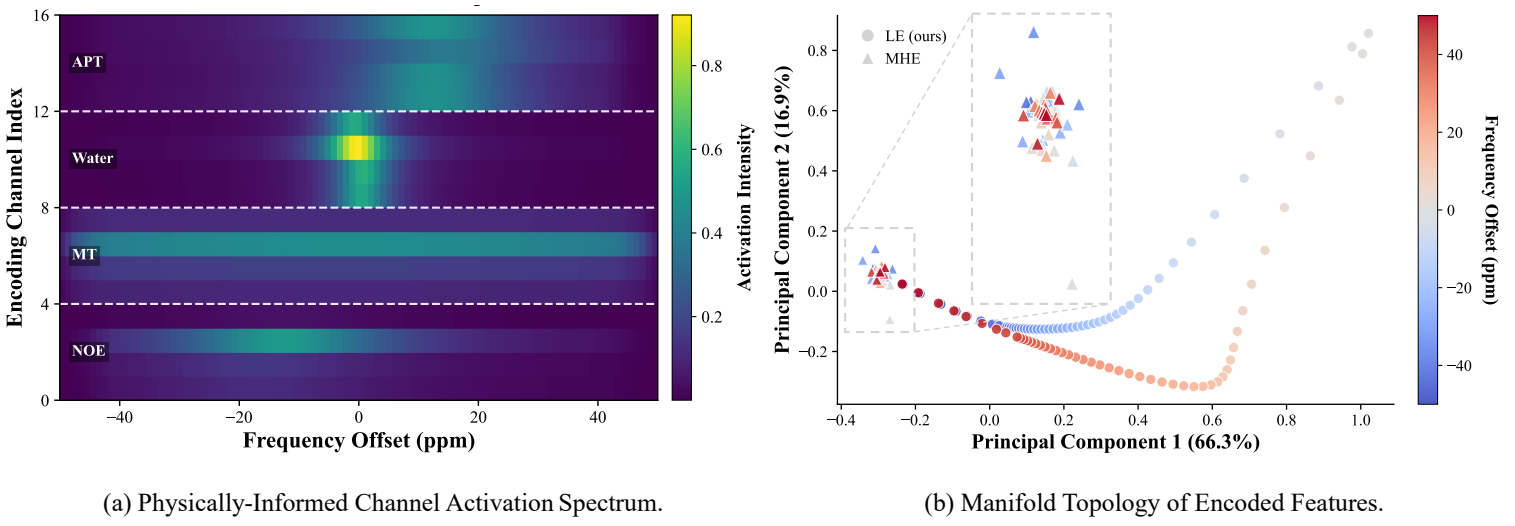

(a) Physically-Informed Channel Activation Spectrum.

(b) Manifold Topology of Encoded Features.

**Fig. 5.** Visualization of the Lorentz Encoding mechanism.

# 4 Discussion

## 4.1 Physical Interpretability and Physics-Informed Regularization

The core innovation of LE lies in transforming the reconstruction problem from unconstrained regression to physics-informed bounded optimization. This mechanism explains its superior performance observed in our results.

As shown in **Fig. 4**, the U-Net method exhibits a clear overfitting behavior. Although it accurately fits the sampled points, it fails to interpolate correctly in gap regions, leading to a flat or erratic response. Similarly, generic INRs (e.g., MHE, WIRE) suffer from severe performance degradation in spectral regions with large sampling intervals (e.g., >10 ppm). Because they operate in an unconstrained high-dimensional space, these models lack the inductive bias to bridge wide spectral gaps, often producing physically invalid oscillations or signal drops that propagate into erroneous parameter maps. In contrast, LE restricts the solution space to a linear combination of Lorentzian functions. Even in the highly under-sampled >10 ppm region, LE maintains high PSNR. This is because the network is forced to follow the smooth, physics-governed curvature of the Lorentzian shapes. The network effectively recovers the correct spectral shape based on physical prior rules rather than blind local interpolation.

This can be also seen from the visualization of the topology of the learned features, as shown in **Fig. 5b**. The features generated by the generic encoding method (MHE) are densely clustered and highly overlapping, failing to capture the relationship between signals of different frequency offsets. In contrast, the LE features form a smooth, continuous V-shaped trajectory in the latent space, with the frequency offsets (represented by the color gradient) smoothly transitioning along the curve. This geometric continuity confirms that the network has successfully learned a regularized physical structure. Furthermore, the channel activation heatmap (**Fig. 5a**) reveals distinct bands corresponding to specific metabolite pools (Water, APT, NOE, MT), demonstrating that LE automatically disentangles the Z-spectra into physically meaningful components without explicit supervision.

## 4.2 Limitations and Future Work

Despite its high reconstruction accuracy, the current framework has two primary limitations. First, training remains computationally intensive. Since our INR model is optimized in a slice-by-slice method, the total training time scales linearly with the volume size. On a standard workstation (NVIDIA RTX 5080), optimizing a single slice requires approximately 5 minutes, making whole-brain reconstruction (typically 20-50 slices) considerably slower than inference-only deep learning methods. Second, while the physical constraint effectively suppresses noise, it can occasionally lead to over-smoothing of fine spatial details. This trade-off is inherent to the regularization strength; strong spectral constraints prioritize spectral continuity, potentially at the cost of high-frequency spatial texture in regions with subtle anatomical variations.

Our future research will focus on integrating the parameter fitting step directly into the optimization loop for end-to-end metabolite mapping like Chen et al.[7]. Furthermore, we plan to extend this framework to joint MRI sequence design—co-optimizing acquisition parameters with our reconstruction network—to maximize efficiency for real-world clinical applications.

## 5 Conclusion

In this work, we proposed Lorentz Encoding (LE), a physics-informed implicit neural representation framework for high-fidelity CEST Z-spectra reconstruction from sparsely sampled data. By embedding Lorentzian priors into the encoding layer and transforming the task into a physical-constrained inverse problem, our method effectively overcomes the limitations of data sparsity and signal saturation. Extensive experiments demonstrate that LE significantly outperforms existing deep learning baselines in both image reconstruction quality and the accuracy of derived clinical parameter maps (APT, NOE, MT). This approach offers a promising solution for accelerating clinical CEST imaging without compromising diagnostic reliability.

**Acknowledgments.** This work was supported by Grand NeoBay Development Fund (202509-KW-1.1.1-001).

**Disclosure of Interests.** The authors have no competing interests to declare that are relevant to the content of this article.